%% file: CVPR_main.tex
\documentclass[10pt,twocolumn,letterpaper]{article}

\usepackage{cvpr}              
\usepackage{graphicx}
\usepackage{multicol}
\usepackage{float}
\usepackage{svg}
\usepackage{balance}
\usepackage{graphicx}

\newcommand{\quotes}[1]{``#1''}


\definecolor{cvprblue}{rgb}{0.21,0.49,0.74}
\usepackage[pagebackref,breaklinks,colorlinks,allcolors=cvprblue]{hyperref}

\def\paperID{00835} 
\def\confName{CVPR}
\def\confYear{2025}

\title{The Art of Deception: Color Visual Illusions and Diffusion Models}

\author{Alex Gomez-Villa$^{1,2}$, Kai Wang$^{1,2}$, Alejandro C. Parraga$^{1,2}$, Bartolomiej Twardowski$^{1,2,4}$,\\ Jesus Malo$^{3}$, Javier Vazquez-Corral$^{1,2}$, Joost  van de Weijer$^{1,2}$\\
$^1$Computer Vision Center, Barcelona, Spain\\
$^2$Universitat Autonoma de Barcelona, Barcelona, Spain\\
$^3$Universitat  de Valencia, Valencia, Spain\\
$^4$IDEAS NCBR, Warsaw, Poland\\
\url{https://alviur.github.io/color-illusion-diffusion}
}

\begin{document}

\twocolumn[{%
\renewcommand\twocolumn[1][]{#1}%
\maketitle
\begin{center}
    \centering
    \captionsetup{type=figure}    
    \includegraphics[width=0.95\textwidth]{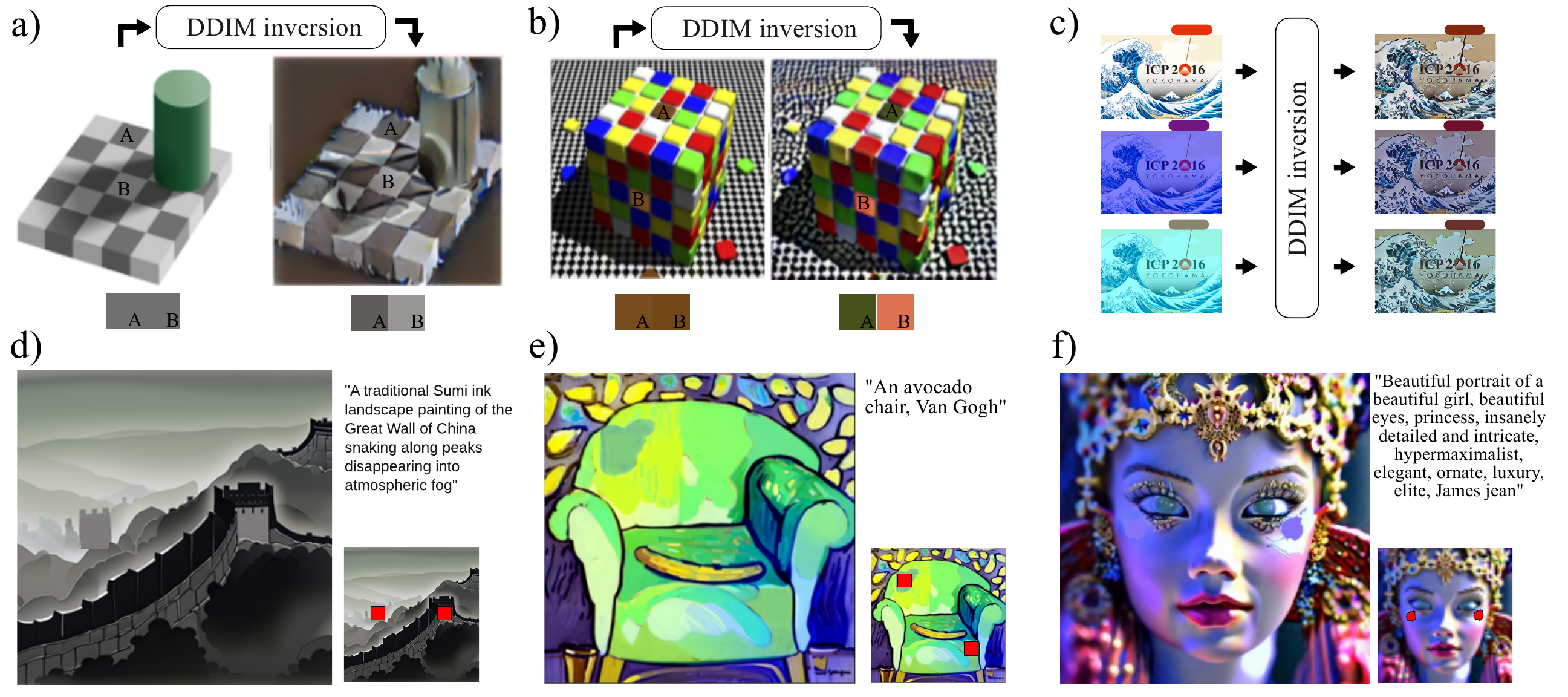}
    \vspace{-3mm}
    \captionof{figure}
    {
    \textbf{Observation: Denoising Diffusion Implicit Models (DDIM) have human-like visual illusions. Application: text-to-image generated visual illusions.}
    Examples (a) and (b) show how the responses to physically equal patches shift differently in the path to the latent space. In contrast, (c) shows how physically different stimuli get more similar along the path, achieving color constancy.
    The observation of this human-like behavior allows us to propose the use of text-to-image models to generate images in which physically identical patches are perceived differently (examples d, e, f). We highly recommend to \textbf{watch the illusions on a computer screen}.}\label{fig:teaser}
\end{center} 
}]

\begin{abstract}
Visual illusions in humans arise when interpreting out-of-distribution stimuli: if the observer is adapted to certain statistics, perception of outliers deviates from reality.
Recent studies have shown that artificial neural networks (ANNs) can also be deceived by visual illusions.
This revelation raises profound questions about the nature of visual information. Why are two independent systems, both human brains and ANNs, susceptible to the same illusions? Should any ANN be capable of perceiving visual illusions? Are these perceptions a feature or a flaw?
In this work, we study how visual illusions are encoded in diffusion models. 
Remarkably, we show that they present human-like brightness/color shifts in their latent space.
We use this fact to demonstrate that diffusion models can predict visual illusions. 
Furthermore, we also show how to generate new unseen visual illusions in realistic images using text-to-image diffusion models. 
We validate this ability through psychophysical experiments that show how our model-generated illusions also fool humans.

\end{abstract}

\vspace{-0.5cm}
\section{Introduction}
\label{sec:intro}
Visual illusions (VIs) serve as a fascinating window into the complex workings of human visual perception, where our interpretation of reality diverges significantly from physical measurements. These phenomena can have a profound social impact, as demonstrated by viral phenomena such as \quotes{The Dress} \cite{gegenfurtner2015many}, which sparked global discussions on the nature of perception. Beyond their intrinsic scientific value, VIs offer a unique lens through which we can examine and compare human and machine perception \cite{gomez2020color,lonnqvist2021comparative}.
The emergence of diffusion models (DMs) as state-of-the-art generative models has revolutionized the field of artificial intelligence, demonstrating unprecedented capabilities in image synthesis and manipulation. Their denoising-based architecture, which gradually transforms random noise into coherent images, presents a novel paradigm for studying visual processing that mirrors the lossy and noisy process of human perception. This parallel makes them particularly interesting for investigating VIs, as both systems must navigate the complex relationship between image statistics and perceptual interpretation.

The study of VIs in artificial neural networks (ANNs) has a rich history. Early work demonstrated that a multilayer perceptron could replicate brightness illusions~\cite{corney2007lightness}, while subsequent research by Gomez-Villa et al. \cite{gomez2019convolutional} showed that convolutional neural networks (CNNs) could mirror human perception in lightness and color illusions. Concurrent developments included Watanabe et al.'s \cite{watanabe2018illusory} demonstration of motion VI replication in video-prediction CNNs and the publication of the first visual illusion dataset \cite{williams2018optical}. This foundation has spawned diverse investigations into the ANN-VI relationship in multiple domains: image completion \cite{kim2019neural,sun2021imagenet}, brightness/color perception \cite{gomez2020color,hirsch2020color,kubota2021machine,li2022visual,lonnqvist2021comparative,ulucan2023investigating}, contrast sensitivity~\cite{li2022contrast,akbarinia2023contrast}, shape perception \cite{mummadi2021does,zhangmuller}, and vision-language models \cite{burgert2023diffusion_illusion,ngo2023clip,shahgir2024illusionvqa,zhang2023grounding}.

In this work, we investigate how human-like perceptual phenomena naturally emerge in the encoding process of DMs. 
Our key finding reveals that the intermediate steps of diffusion models exhibit trends that align remarkably with human perception. Specifically, when analyzing the trajectory of an inverted image through the diffusion process, we observe that the progressive changes in lightness and color properties closely mirror human perceptual responses.
Prior research has demonstrated that Gaussianization flows exhibit interesting properties when processing samples that deviate from their training data statistics. When trained on specific sample distributions, these flows can produce unusual representations for out-of-distribution samples~\cite{Rezende15,Laparra11} - a characteristic that has proven valuable for measuring distribution divergence~\cite{Laparra24}. Intriguingly, this phenomenon parallels human visual perception, where shifts in image statistics can lead to illusory perception of spatio-chromatic patterns, a behavior that has been explained through manifold equalization~\cite{Laparra12,Laparra15}.
We argue that image samples that do not follow the statistics of natural images are not properly gaussianized by DMs. In this way, 
linking defective gaussianization and illusory perception of unusual samples, it makes sense that
intermediate representations in DMs may be correlated with the shifts of these samples for human observers.
The reported alignment suggests that the iterative denoising in diffusion models may inadvertently capture fundamental aspects of human visual perception, particularly in contexts where our perception diverges from physical reality. By examining these intermediate representations, we provide novel insights into the nature of visual illusions and the unexpected parallels between human perception and the diffusion process.

\noindent \textbf{Contributions:} The main contributions of this work are:
\begin{itemize}[leftmargin=*]
    \item We report an empirical observation of the behavior of diffusion models: progressive changes in lightness, hue and saturation of image regions in the forward/inverse transform to the latent representation replicate the shifts of human perception of these regions.
    \item We derive a novel approach to using diffusion models to model human vision. We show that deep autoencoders can be effectively employed as vision models by performing a diffusion inversion process and measuring perception in the intermediate noise space.
    \item We demonstrate the human perception replication capabilities of DM in many cases, including VI datasets, popular published VI works, and natural images.
    \item Leveraging the perceptual capabilities of diffusion models, we propose a novel method for generating  brightness/color VIs using text-to-image models. 
\end{itemize}

\section{Related work}
\label{sec:related_work}

\paragraph{Replication of Visual Illusions.}
Replication of visual illusions has been made with several methods, but all of them can be roughly classified into the following categories: 
\begin{itemize}[leftmargin=*]
    \item Image restoration methods~\cite{gomez2019convolutional,gomez2020color,kubota2021machine,li2022contrast,lonnqvist2021comparative}: These use pre-trained image restoration backbones. They process VIs and measure the output image to measure perception.
    \item Effect bias methods~\cite{akbarinia2023contrast}: These works are inspired by neuroscience measurement of stimuli. Pretrained architectures are used to process VIs, then the response in the internal activations of the ANN is used to define if there is a visual illusion.
    \item Likelihood method~\cite{hirsch2020color}: In this approach, the most probable perceived intensity of each pixel in an image is estimated using an invertible flow trained on natural images.
    \item Language-guided methods~\cite{ngo2023clip,zhang2023grounding}: Language is used as a vehicle to measure perception. These works use vision-language models as test subjects and ask them questions regarding the perception of the image.
\end{itemize}

Our approach differs from previous methods in the following: 1) we do not measure perception in the output of the model. Instead, we measure the VI in the intermediate steps of image inversion process. 2) We use a deep autoencoder; previous approaches in this category claim that deep autoencoders suffer from less perceptual bias and use shallow architectures for replication experiments. Furthermore, we are able to replicate human perception in high resolution images (as big as DMs are able to manage) and in natural images. These setups were not possible in previous works.

\paragraph{Generation of Visual Illusions.}
Although generative models have been used to synthesize visual illusions~\cite{gomez2022synthesis}, DMs has only recently been proposed as the generative engine. In ~\cite{burgert2023diffusion_illusion} high-level illusions (overlay effects) are generated by mixing a set of images  and transformations together. Later, ~\cite{geng2024factorized_diffusion,geng2023visual_anagrams}, were able to create more general illusions such as rotations, color inversions, and hybrid images.

The work most similar to ours is by Roy et al.~\cite{roy2024bri3l}, who used stable diffusion (SD) to generate brightness visual illusions through keyword prompts. However, their approach neither explicitly optimizes perceptual effects nor specifies regions for perception replication. Consequently, illusions may appear randomly in the generated image with unknown localization. Additionally, their generated images are limited to conventional brightness illusions (gratings or high-frequency black and white patterns) and cannot incorporate visual illusions into natural images.
In contrast, our work allows visual illusions within generated natural images as well as in typical high-frequency patterns. We explicitly generate the perceptual effect in a designated image region and can control the image content through text prompts.

\section{Methods}

\subsection{Diffusion models}

Diffusion models (DMs) are generative models that learn to gradually denoise images by reversing a fixed Markov chain of forward diffusion steps. The forward process systematically adds Gaussian noise to an image until it becomes pure noise, while the learned reverse process reconstructs the original image through iterative denoising. While traditional diffusion follows a stochastic sampling process, Denoising Diffusion Implicit Models (DDIM) \cite{song2021ddim} introduce a deterministic sampling approach that maintains high  quality while enabling significantly fewer inference steps.

DDIM's deterministic nature enables a crucial capability: the ability to invert generated images back into their latent representations through a process known as DDIM inversion \cite{song2021ddim}. This inversion process allows us to map any image into the model's latent space and observe its trajectory through the diffusion process. Given an image $z_0$, DDIM inversion produces a sequence of intermediate representations $\{z_t\}_{t=1}^T$ according to:
\begin{equation}
\begin{split}
z_t = \sqrt{\bar{\alpha}_t} \, z_0 
    + \sqrt{1-\bar{\alpha}_t} \, \epsilon_{\theta}^{z_t,t}
\end{split}
\end{equation}
where $\bar{\alpha}_t$ represents the cumulative product of noise scheduling coefficients, and $\epsilon_{\theta}^{z_t,t}=\epsilon_\theta(z_t,t)$ is the learned noise prediction network. The reverse process follows:
\vspace{-0.05cm}
\begin{equation}
\hspace{-0.4cm} z_{t-1}  =  \sqrt{\bar{\alpha}_{t-1}}\left(\frac{z_t - \sqrt{1-\bar{\alpha}_t} \, \epsilon_{\theta}^{z_t,t}}{\sqrt{\bar{\alpha}_t}}\right) + \sqrt{1-\bar{\alpha}_{t-1}} \, \epsilon_{\theta}^{z_t,t}
\end{equation}

DDIM inversion has emerged as a powerful tool across various image manipulation tasks. Its ability to faithfully reconstruct latent representations has enabled applications such as semantic image editing \cite{meng2022sdedit}, style transfer \cite{kwon2023diffusionbased}, image interpolation between two sources \cite{wang2023interpolating}, and controlled image generation through classifier-free guidance \cite{ho2022classifier}. In our work, we exploit DDIM inversion's deterministic nature to systematically study how visual illusions are processed through the diffusion trajectory.

\subsection{DDIM inversion may replicate human vision}
\label{sec:replication}

The \textit{wholly-empirical evolutionary paradigm}~\cite{kingdom2011lightness} suggests that human vision is shaped by statistical regularities in natural scenes, where our visual system  evolved to interpret ambiguous stimuli based on their likelihood. 
We leverage this insight to study how diffusion models, trained for natural images, process illusory stimuli, which \emph{are not} in the distribution of natural images.

\begin{figure*}[t]
    \centering
    \includegraphics[width=0.9\textwidth]{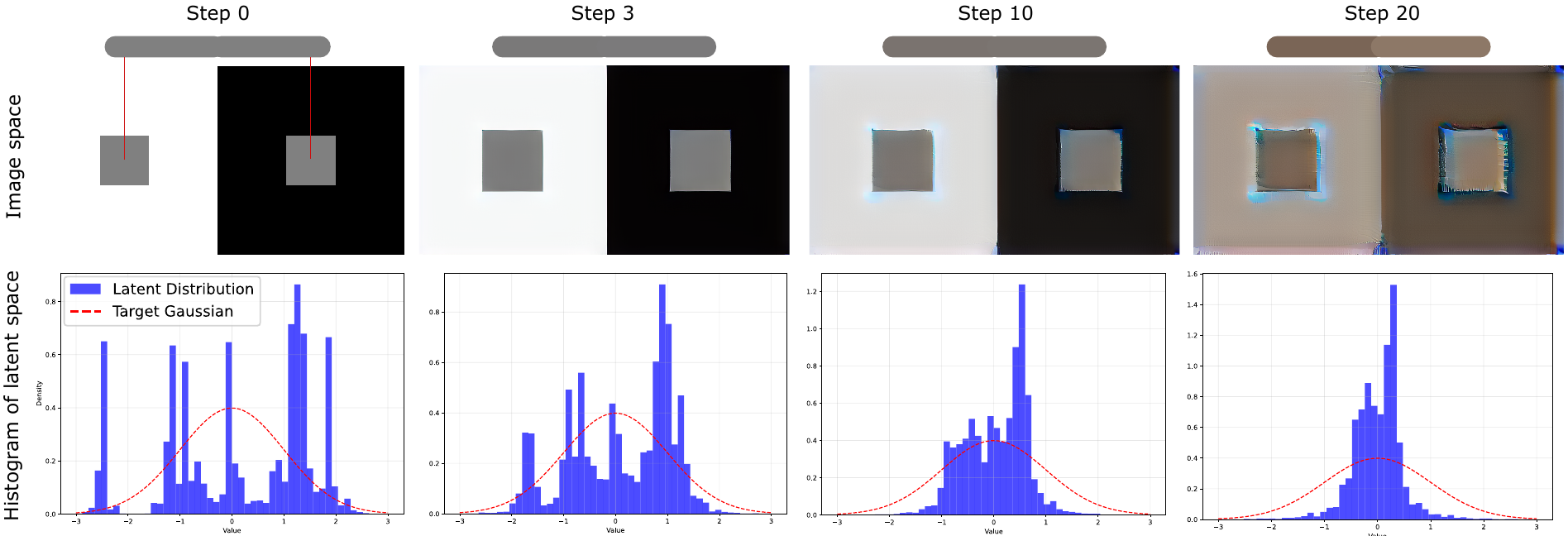}
    \vspace{-3mm}
    \caption{\textbf{DDIM inversion of the Brightness Contrast illusion~\cite{bruke} using Stable Diffusion}. Top row: Image-space visualization (decoded latents) showing (left) Original illusion with two identical gray squares (marked in red) against black and white backgrounds, and inversion results using 3, 10, and 20 steps. Bottom row: Histograms of the corresponding latent representations. The model gradually reproduces the perceptual difference in brightness between the physically identical squares in a (not fully Gaussian) intermediate representation.}
    \vspace{-2mm} 
    \label{fig:main}
\end{figure*}

\noindent \textbf{Motivation.}
In Fig.~\ref{fig:main}, we present an illustrative experiment using Stable Diffusion~\cite{rombach2022high} (extra results with an image space model in the supplementary material). We perform DDIM inversion on the canonical visual illusion known as Brightness Contrast~\cite{bruke}. 
DDIM inversion puts the image back into the latent space.
This illusion presents two gray squares of identical intensity (indicated by red markers) against white and black backgrounds, which leads humans to perceive the left square as darker than the right one. 
The first row  (left to right)
shows the inversion\footnote{we decode the latent representation to visualize it in image space.} for $0$, $3$, $10$, and $20$ steps, respectively. As the zoomed-in bar shows, the initially equal-intensity regions progressively develop an intensity difference that aligns with human perception\footnote{In addition to the color/brightness effects, notice how the high-frequency content of the image is maintained; we explore this phenomenon further in supplementary material.} (left target appearing darker than right). 
The second row shows the progressive Gaussianization of the latent representation. For this image perception is replicated at an intermediate representation where gaussianization is not complete (since this stimulus is not in the distribution of natural images).

\noindent \textbf{Observation: emergence of perceptual biases.}
Our key observation is: \textit{when inverting an image through diffusion models the intermediate denoising steps progressively introduce color/brightness shifts that align with human perception}. We hypothesize that this occurs because diffusion models, trained on natural images, map illusory stimuli onto the learned natural image manifold, resulting in systematic “interpretation errors” that mirror human perceptual biases.

\noindent \textbf{Theoretical framework.}
Formally, let $G: \mathcal{X} \rightarrow \mathcal{Z}$ be the forward diffusion mapping from image space $\mathcal{X}$ to latent space $\mathcal{Z}$. For natural images $x \in \mathcal{X}$, $G$ learns a mapping $z = G(x)$ that transforms the image distribution into a Gaussian. However, when presented with an eventually illusory stimulus $y$, we have $y \notin \mathcal{X}$. Still, the mapping $z' = G(y)$ attempts to project the illusion onto the learned natural image manifold, introducing systematic deviations. The intermediate states $z_t$ reveal how the model progressively ``corrects'' the illusion toward natural image statistics. We observe that these corrections correlate strongly with human perception, suggesting that both humans and diffusion models may operate on similar statistical principles when processing ambiguous visual input. This alignment between diffusion trajectories and human perception provides computational evidence for the wholly-empirical paradigm, demonstrating how statistical learning can lead to perceptual illusions without explicit encoding of perceptual rules. The systematic nature of these ``errors'' suggests they emerge from the same statistical regularities that shape human visual processing through evolution.

\subsection{Generating visual illusions with DMs}

Based on the findings above, we develop a method to automatically generate brightness/color illusions with DMs. To do so, we need to evaluate whether a VI is present and then push the diffusion process toward the desired perception.

Previous works on VIs and DMs focus on high-level visual illusions, i.e. cognitive effects that measure if the image belongs to a certain class of prompt~\cite{burgert2023diffusion_illusion,geng2023visual_anagrams,geng2024factorized_diffusion}. Here we focus on brightness/color VIs, which leads to the challenge of measuring effects in specific regions of the image. 

Here we answer the measurement questions proposed in previous works on brightness/color VI generation in this way:

\noindent \textbf{Where to measure?} This question separates the generation of high-level VIs from color/brightness VIs. Models aiming at high-level VIs that fool an ANN require to ask a classifier/vision-language model which class/prompt better describes the image~\cite{geng2023visual_anagrams, ngo2023clip,zhang2023grounding}. In contrast, color VIs appear in specific regions of the image, therefore, they need to be measured at those points. In this work, we specifically define regions (here \textit{targets}) in which we want a perceptual effect to appear. These targets are our measurement points, and we guarantee that in these regions, we can track, measure, and optimize perceptual effects. Failing to define a target leads to failure, as reported in~\cite{williams2018optical}.

\noindent \textbf{What to measure?} During image generation, we will measure the intensity values of the intermediate steps $z_t$. Even though $z_t$ does not contain the final perception bias ($z_T$), we can push the generative process towards a desired perceptual effect using small changes in the intermediate representation as explained in Fig.~\ref{fig:how_to}. 

\noindent \textbf{How to measure?} In the motivating example of section~\ref{sec:replication} (Fig.~\ref{fig:main}), the model replicates human perception if the target shifts in the same direction as humans perceive. However, in an optimization procedure, we need to measure how "good" the illusion is; in other words, we need to quantify the effect. Since we are concerned about brightness and color VI, we will measure the mean intensity per channel of each target. More precisely, given an image $I$, and the target region $r \in \mathbb{R}^2$ in $I$, we compute the mean intensity in that target region as:
\begin{equation}\label{eq:meter_gray}
    m_{int}(I,r)= \frac{1}{M} \sum_{i\in r} I(r_i)
\end{equation}
where $r_i$ denotes a specific pixel $i$ in target region $r$, $M$ stands for the number of pixels in the target region, and $m_{int}$ is measured in each channel. Hence, the perceived intensity is an RGB triplet ($m_{int}^{R},m_{int}^{G},m_{int}^{B}$).

\subsubsection{Visual perception inside generation pipeline.} 
\begin{figure}[tb]
    \centering
    \includegraphics[width=0.48\textwidth]{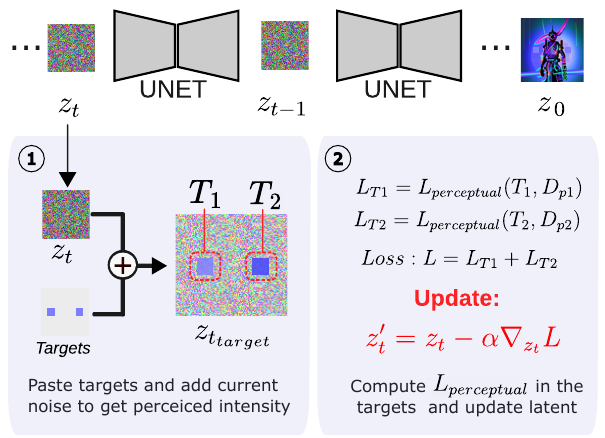}
    \vspace{-4mm}
    \caption{\textbf{Overview of our visual illusion generation pipeline.} The process modifies noisy latent representations, $z_t$, through a custom loss function, $\mathcal{L}$, that guides the generation toward perceptually ambiguous outputs. }
    \vspace{-2mm} 
    \label{fig:how_to}
\end{figure}

Since text-to-image (T2I) diffusion models often struggle to perfectly align with textual prompts, a variety of augmentation methods have been developed to enhance the fidelity of T2I generation. These methods typically focus on optimizing noisy latent representations in each timestep, a strategy that has proven effective in refining generation outcomes~\cite{chefer2023attend, rassin2024linguistic_binding, wang2023tokencompose, zhang2024enhancing}. A common approach involves the design of an update loss function, $\mathcal{L}$, which is tailored based on the layout of the noisy latent or the characteristics of the cross-attention maps. 
This loss is then used to guide the backpropagation process, updating the noisy latent in a manner that more accurately reflects the intended textual descriptions, thereby enhancing the overall quality of the generated images. This process can be formulated as $z^{'}_t = z_t - \alpha \cdot \nabla_{z_t}\mathcal{L}$.
In this paper, we diverge from the conventional use of optimization techniques aimed at aligning T2I diffusion models with textual semantics. Instead, we leverage these techniques to create visual perception effects (see Fig.~\ref{fig:how_to}). We use an image space DM (DeepFloyd~\cite{DeepFloydIF}) to avoid the use of the decoder in the backpropagation). 

First, we define the original target's image as $O(r)$, where $r \in \mathbb{R}^2$. We fix $O(r)$ to have the original value given to the targets by the user at each color channel. Then, given an intermediate image $z_t$, and a list of $R$ target regions $\{r^p\}_{p=1}^R$, 
our process to generate a VI includes the addition of the targets to the image in the desired region\footnote{Note that this addition results in the perceived color (see sect. 3.2)}, i.e. we compute $z_{t_{target}}(r_i^p)=z_t(r_i^p)+O(r_i^p)$, for all pixels $i$ inside the region $r^p$ and $z_{t_{target}}=z_t$ for any pixel outside this target region. Finally, the following optimization objective is applied $N$ times on each diffusion step $t$ to guide the generation: 
\begin{equation}\label{eq:loss_function}
    \mathcal{L} = \gamma \mathcal{L}_{VI} +  \beta \mathcal{L}_{sim}
\end{equation}

where $\mathcal{L}_{VI}$ and  $\mathcal{L}_{sim}$ measure the illusion strength 
and the similarity between targets and background respectively, and 
$\gamma$ and $\beta$ are hyperparameters that control the relative weight of these measures. Specifically:

\begin{itemize}

\item $\mathcal{L}_{VI}$ is a perceptual loss on the target intensities: 
\begin{equation}\label{eq:loss_VI}
    \mathcal{L}_{{VI}} = \sum_{c \in R,G,B}\sum_{r^p}^R \mathcal{L}_{Percept}(z_{t_{target}},k_{c}^p;r^p)
\end{equation}
where $r^p$ is a specific target, $k_c^p$ is the desired intensity for target $r^p$ on color channel $c$, and
\begin{equation}\label{eq:loss_Perc}
    \mathcal{L}_{Percept}(z_{t_{target}},k_c;r)=\|m_{int}(z_{t_{target}},r)-k_c\|
\end{equation}
with $m_{int}$ as defined in Eq. \ref{eq:meter_gray}.

\item $\mathcal{L}_{sim}$ is a similarity term. This term aims at avoiding the region from appearing ``out of place''. In particular, it looks at harmonizing the targets with the background-generated image:  
\begin{equation}
    \label{eq:harmonize}
    \mathcal{L}_{sim} = \sum_{p = 1}^R MSE( \, z_{t}(r^p),O(r^p) \, )
\end{equation}

\end{itemize}

\section{Experimental setup}

\subsection{Datasets}

\begin{itemize}

    \item \textbf{Brightness illusions dataset}: The \textit{BRI3L}~dataset~\cite{roy2023bri3l} consists of $22366$ grayscale images spanning five types of brightness illusions. Each image is annotated with binary masks indicating regions of illusory perception. 
    \item \textbf{Visual Illusion VQA Datasets}: Several datasets have been developed to evaluate vision-language models' understanding of VIs: \textit{IllusionVQA}~\cite{shahgir2024illusionvqa} contains 435 illusory images with multiple-choice questions across 12 categories, plus 1000 impossible object image pairs for localization testing. \textit{GVIL}~\cite{zhang2023grounding} features 100 base illusion images expanded to 1600 instances covering five illusion categories, with four evaluation tasks: same-difference QA, referential QA, attribute QA, and referential localization. \textit{HallusionBench}~\cite{guan2024hallusionbench} comprises 346 images and 1129 questions designed to test image-context reasoning and diagnose both VIs and language hallucinations. We use the color category in all the VQA datasets.
\end{itemize}

\subsection{Benchmarks}

\begin{itemize}
    \item \textbf{Classical vision science models:} These approaches are based in vision science and need to be calibrated for each image to fully exploit their capabilities. In this paper we use the parameters set by Betz et al.~\cite{betz2015noise}. These parameters constitute a fair guideline for these methods, as it is not possible to fine-tune them for thousands of images. We use ODOG (Oriented Difference of Gaussians)~\cite{blakeslee1999multiscale} and the Chromatic Induction Wavelet Model (CIWaM)~\cite{otazu2010toward}
    \item \textbf{Neural networks-base models:} We compare against previous image restoration methods that proven ability in visual illusion detection: DN-NET\cite{gomez2020color} and Restorenet\cite{li2022contrast}. 
    \item \textbf{Variational autoencoder (VAE):} We present the results of only considering the VAE of SD prior to the DDIM inversion process ---to further demonstrate DDIM is the responsable for the alignment with human perception.
\end{itemize}

\subsection{Metrics}

While visual illusion replication results have traditionally been presented through profile plots comparing model outputs with human perception along image cross-sections, we leverage recent datasets to provide quantitative metrics. This approach complements classical visualization methods (provided in supplementary materials) with rigorous numerical evaluation.

For the BRI3L dataset, we compute the mean intensity difference, $\Delta I$, between the model's output and the input image within each segmented region:
\begin{equation}
\Delta I = \frac{1}{|M|} \sum_{i \in r} |I_{out}(r) - I_{in}(r)|
\end{equation}
where $M$ represents the size of the segmentation mask, and $I_{out}$, $I_{in}$ are the output and input intensities respectively. We consider the model's response to be aligned with human perception when $\Delta I < \tau*I_{int}$, where $I_{int}$ represents the mean intensity of the input image in the target region (calculated using Eq.~\ref{eq:meter_gray}). Since BRI3L aims to detect regions that become darker, the metric becomes more harder as the value of $\tau$ decreases, requiring the model to achieve greater darkness in the target regions.

For the VQA-based datasets (IllusionVQA, GVIL, and HallusionBench), we manually create segmentation masks and consider the answer of the model aligned with human perception if it matches the expected intensity for regions referenced in the question-answer pairs.

What we refer to as Perception Accuracy Score (PAS) is the percentage of images having the correct perceptual predictions in each dataset.

\subsection{Experiments}

We conducted two sets of experiments to evaluate our method's ability to: (1) replicate existing VIs across multiple datasets, and (2) generate novel VIs that can fool human perception.

For each experiment we present qualitative results zooming in the region of the perceptual effect and quantitave evaluations using the proposed metrics.

\begin{enumerate}
    \item \textbf{Replication of visual illusions:} We test our model's ability to replicate brightness (BRI3L) and color (IllusionVQA, GVIL, and HallusionBench) illusions by comparing the intensity differences between input and output images within annotated segmentation masks. For BRI3L, we evaluate performance across three $\tau$ levels (0.8, 0.9, 1.0) to assess the robustness of illusion replication. In this experiment we present results using SD 1.4 (latent space DM) but additional results with DeepFloyd (image space DM) are provided in the supplementary material.

    \item \textbf{Generation of visual illusions:} In this experiment, we check whether we can generate VIs in images that fall inside the distribution of the T2I model. We use guidance scale~$=10$ (we want to generate images that follow the prompt),$N=5$, $\gamma = 0.5$, and $\beta = 1.0$. Additionally, here we use S$2$ of DeepFloyd to generate higher-resolution versions of our generated visual illusions. For flat intensity targets (non-gradients), we intentionally paste the \textit{targets} in the output of S$2$ as the superresolution model may change the content of the targets. 

\end{enumerate}

\section{Results}
\subsection{Replication of visual illusions}

\paragraph{Qualitative results.}
\begin{figure*}[ht]
    \centering
    \includegraphics[width=0.98\textwidth]{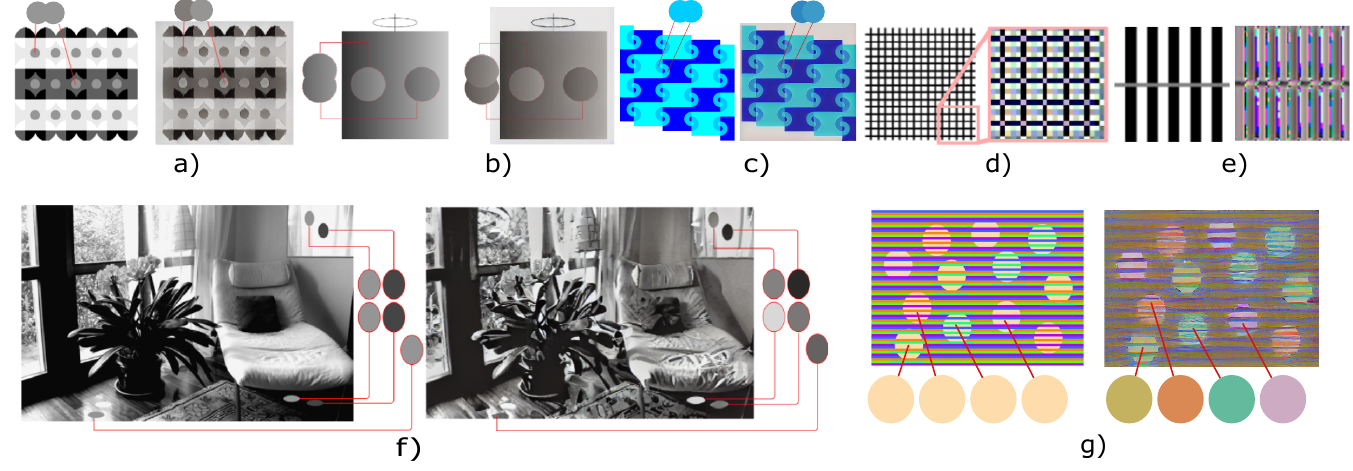}
    \vspace{-4mm}
    \caption{\textbf{Qualitative replication of visual illusions (please enlarge display)}: For each item, the original image is on the left and its DDIM inversion on the right. The observation done in Fig.~\ref{fig:main} is consistently reproduced. We zoom in on the region containing the illusion in each image pair. a) Barutan-seijin~\cite{barutan}, b) Robot~\cite{robot}, c) Shiosai~\cite{shiosai}, d) Herman-grid~\cite{hermann1870erscheinung},  e) Grating induction~\cite{mccourt1994grating}, f) Bright room~\cite{natural}, g) Confetti illusion~\cite{confetti}. For more qualitative examples and results of classical visual illusions, see the supplementary material.}
    \vspace{-2mm} 
    \label{fig:rep_quali}
\end{figure*}

We present qualitative results demonstrating our method's ability to replicate human visual perception across diverse scenarios in Fig.~\ref{fig:rep_quali}. Each comparison shows the original input image (left) and its corresponding DDIM inversion after 5 steps (right), with zoomed inset regions highlighting the areas where perceptual effects occur. The top row showcases our method's effectiveness on classical visual illusions, including Hermann grid and grating induction patterns. The bottom panels demonstrate the model's capability to capture perceptual phenomena in both natural grayscale photographs and artistic images with different color schemes. The zoomed regions, emphasize physically identical targets that are perceived differently due to their surrounding context — a key characteristic of visual illusions that our model successfully replicates. Additional qualitative results showing our method's performance across a broader range of visual illusions and natural images can be found in the supplementary material.

\paragraph{Quantitative results.}

\begin{table}[b]
\centering
\caption{\textbf{Perceptual Accuracy Score (PAS) in BRI3L for different methods and thresholds}}
\vspace{-2mm}
\begin{tabular}{lccc}
\toprule
Method / Threshold & 0.8 & 0.9 & 1.0 \\
\midrule
ODOG~\cite{blakeslee1999multiscale} & 36.58 & 57.09 & 72.44 \\
CIWAM~\cite{otazu2010toward} & 5.69 & 18.23 & 56.22 \\
\midrule
DNnet~\cite{gomez2019convolutional} & 0.00 & 0.00 & 71.09 \\
RestoreNet~\cite{gomez2020color} & 0.00 & 0.00 & 74.95 \\
\midrule
VAE & 0.00 & 0.08 & 73.18 \\
DDIM 5 steps & 44.02 & 91.43 & 99.88 \\
DDIM 10 steps & \textbf{84.17} & \textbf{98.85} &\textbf{ 100.00} \\
\bottomrule
\end{tabular}
\vspace{-2mm}
\label{tab:comparison}
\end{table}

\begin{table}[t]

\centering
\caption{\textbf{Perceptual Accuracy Score (PAS) in different VI sets}}
\vspace{-2mm}
\resizebox{\linewidth}{!}{
\begin{tabular}{lccc}
\toprule
Method & IllusionVQA & GVIL & HallusionBench \\
\midrule
CIWAM~\cite{otazu2010toward} & 56.52 & 63.63 & 83.33 \\
\midrule
DNnet~\cite{gomez2019convolutional} & 14.20 & 21.30 & 33.30 \\
RestoreNet~\cite{gomez2020color} & 14.20 & 21.30 & 33.30 \\
\midrule
VAE & 4.70 & 21.30 & 0.00 \\
DDIM 5 steps & 90.40 & 80.30 & 100.00 \\
DDIM 10 steps & \textbf{90.40} & \textbf{96.70} & \textbf{100.00} \\
\bottomrule
\end{tabular}

\vspace{-2mm}
\label{tab:illusion_benchmarks}
}
\end{table}

Table \ref{tab:comparison} presents a comprehensive comparison of our DDIM-based approach against both classical vision models and modern deep learning methods across multiple visual illusion datasets. The results demonstrate the consistency of the observation on diffusion models reported in Section 3.2 over a range of datasets and their superior  performance in replicating illusions.
Classical models like ODOG and CIWAM, while physiologically grounded, show limited performance and flexibility. These methods require careful parameter tuning for each specific type of visual illusion, as evidenced by their poor scores. 
For instance, ODOG can achieve 72.44\% accuracy on BRI3L but cannot generalize to color datasets, while applicacble in color CIWAM shows poor performance.
Previous deep learning approaches 
show consistent but modest performance across different datasets. 
Their uniform performance across sets suggests they capture some general perceptual principles, but their relatively low accuracy indicates limitations in capturing human behavior.
Our proposed DDIM-based method significantly outperforms all baseline approaches across all datasets. 
The substantial performance gap between our DDIM approach and a standard VAE (which shows near-zero performance on most datasets) highlights the crucial role of the diffusion process in capturing perceptual phenomena.
These results demonstrate that our DDIM-based approach not only achieves state-of-the-art performance but also offers superior generalization across different types of visual illusions without requiring illusion-specific parameter tuning. The consistent high performance across all datasets suggests that the diffusion process naturally captures fundamental aspects of human vision.

\begin{figure*}[ht]
    \centering
    \includegraphics[width=0.99\textwidth]{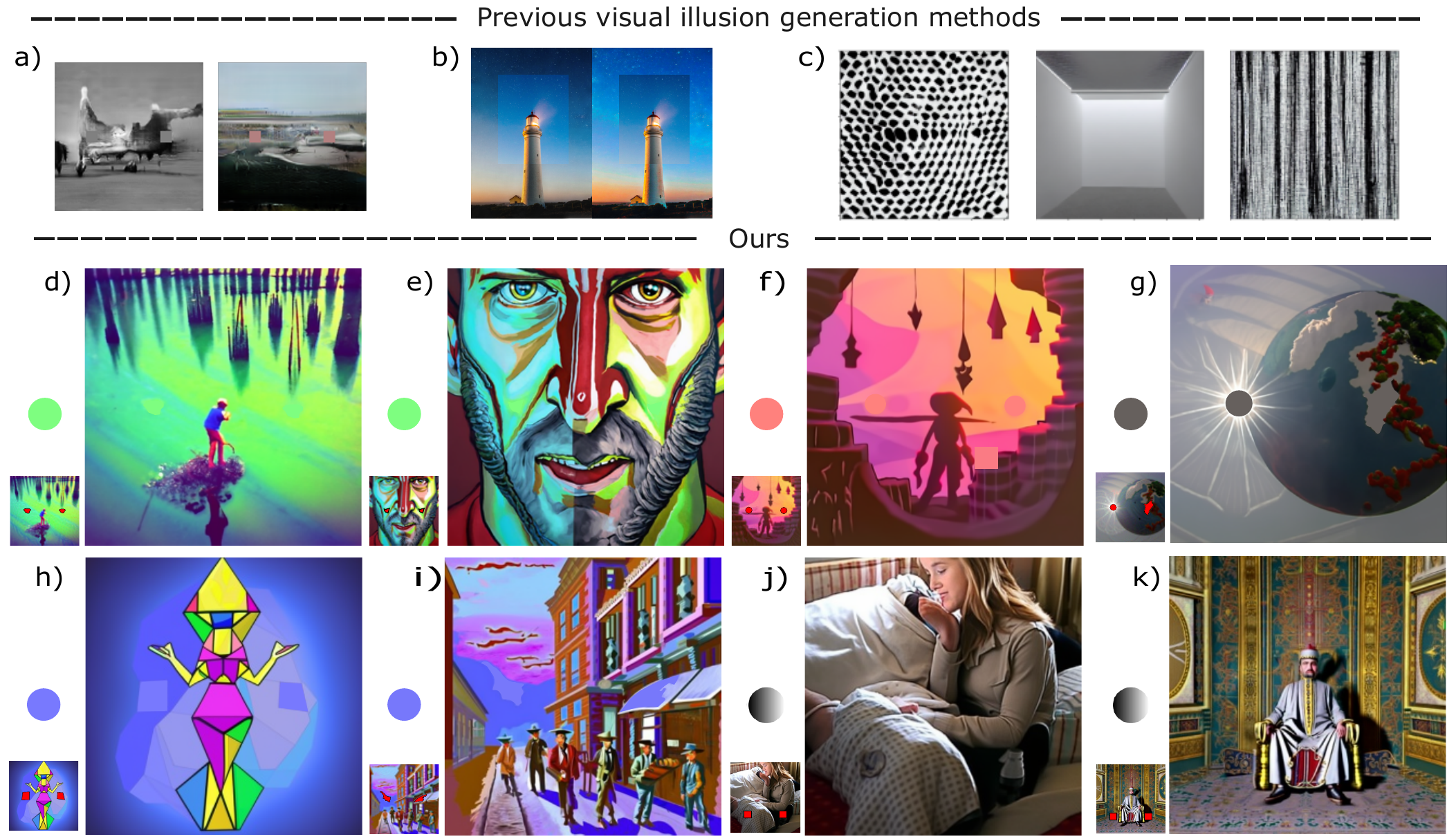}
    \vspace{-2mm}
    \caption{\textbf{Qualitative results of visual illusion generation (viewing at larger scale recommended).} The first row presents prior work by (a) Gomez-Villa et al.~\cite{gomez2022synthesis}, (b) Hirsch et al.~\cite{hirsch2020color}, and (c) Roy et al.~\cite{roy2024bri3l}. The second and third rows display results from our method (text prompts are available in the supplementary material). Images (d) through (i) use identical target colors (shown at the left of each image). In images (j) and (k), we employ grayscale gradients as targets. The selected target regions are highlighted in red in the thumbnail images. 
    }
    \vspace{-3mm}
    \label{fig:sample_natural_generations}
\end{figure*}

\subsection{Generation of Visual Illusions}

\paragraph{Qualitative results.}
Fig.~\ref{fig:sample_natural_generations} demonstrates visual illusions using a 6×6 pixel square target in the S1 stage of DeepFloyd (expanding to 24×24 after S2). The figure displays five different target intensities (shown as circles to the left of each image) with different prompts for each image (selected randomly from DiffusionDB~\cite{wangDiffusionDBLargescalePrompt2022}). Additional results are presented in Fig.~\ref{fig:teaser}. 
In contrast to previous approaches, our method can generate complex content that simultaneously satisfies both the text prompt and the perceptual conditions. To see generation of classic VI (high frequency images with flat patches) please see supplementary material.
\vspace{2mm}

\noindent \textbf{Psychophysical confirmation.} We created $40$ images in which we enforced our model to provide an illusion with either red, green, blue, or achromatic targets. Similarly, we have also created 40 control images—images for which there was no enforcement for the patterns to look different (see Appendix for examples). 15 observers participated in our study. The experiment was conducted in a dark room with controlled dim illumination. The only light source was the monitor, set to sRGB. All observers were tested for correct color vision using the Ishihara test. The screen showed one of the images at a time. Observers were asked to decide whether the two targets were identical or different. The order of images was randomized. We analyze the results in terms of the number of times an image was said to have different targets. For the images in which our model enforced the existence of the illusion, the observers said the targets were different in $64\%$ (mean) and $67\%$ (median) of the cases with a standard deviation of $16\%$. In contrast, only in $13\%$ (mean) and $7\%$ (median) ---standard deviation $14\%$--- of the cases where the control images said to have different targets. 

We also perform an experiment to evaluate the effectiveness of our perceptual loss weight ($\gamma$) through comparison with the two classical models---ODOG and CIWAM--- across different thresholds. As the value $\gamma$ increases, there is a substantial improvement in illusion strength, therefore showing the effectiveness of our perceptual loss. Full details of this experiment are in the Supplementary Material.

\section{Limitations}

\textbf{Replication.} As expected from the image inversion process of diffusion models, the input images are spatially destroyed as the samples make their way to the Gaussian distribution. This implies that the target regions disappear for images closer to the natural image distribution, and the effect can no longer be measured. As a consequence our method has difficulty replicating the visual perception of effects which are based on small regions.

\noindent\textbf{Generation.} Our T2I VI generation method is based on defined regions during inference. This constraints the expressiveness of the T2I model in fulfilling the perceptual conditions. An ideal implementation should establish perceptual regions guided by the attention maps that the input prompt generates to produce more natural visual illusions.

\section{Conclusions}

In this work, we make two key contributions. First, we report and demonstrate over a wide range of datasets an intriguing empirical observation: diffusion models naturally replicate human responses to brightness and color visual illusions through their image inversion process. This remarkable alignment between human perception and diffusion trajectories suggests that both systems may operate on similar statistical principles when processing ambiguous visual input. Second, leveraging this insight, we develop a novel method for generating visual illusions using text-to-image diffusion models, validated through psychophysical experiments. Looking forward, our findings about perceptual processing in diffusion models could help bridge an important gap in current vision-language models, which often struggle with nuanced visual understanding. The way diffusion models progressively process and interpret visual information might offer valuable insights for developing more perceptually-aware foundational models that better align with human visual cognition. This integration of perceptual understanding into image editing workflows represents an exciting direction for future research, potentially leading to tools that naturally account for human visual perception in their operations. 

{
    \small
    \bibliographystyle{ieeenat_fullname}
    \bibliography{longstrings,bibliography}
    \balance

}

\end{document}